\documentclass[letterpaper, 10 pt, conference]{ieeeconf}  

\IEEEoverridecommandlockouts                              

\usepackage{amsmath,amsfonts}
\usepackage{algorithmic}
\usepackage{array}
\usepackage{textcomp}
\usepackage{stfloats}
\usepackage{float}
\usepackage{url}
\usepackage{verbatim}
\usepackage{graphicx}
\def\BibTeX{{\rm B\kern-.05em{\sc i\kern-.025em b}\kern-.08em
    T\kern-.1667em\lower.7ex\hbox{E}\kern-.125emX}}
\usepackage{balance}
\usepackage{graphics} 
\usepackage{wrapfig}
\usepackage{stackengine}
\usepackage{subcaption}
\usepackage{mwe}
\usepackage[british]{babel}
\usepackage{hhline}
\usepackage{multirow}
\usepackage{amssymb}
\usepackage{xcolor,pifont}
\usepackage{tikz}

\usepackage[version=4]{mhchem}

\usepackage{hyperref}
\usepackage{caption}
\usepackage{comment}
\usepackage{booktabs}

\newcommand*\colourcheck[1]{%
  \expandafter\newcommand\csname #1check\endcsname{\textcolor{#1}{\ding{52}}}%
  }
\colourcheck{blue}
\colourcheck{green}
\colourcheck{red}
\definecolor{mypink1}{rgb}{0.858, 0.188, 0.478}
\definecolor{mypink2}{RGB}{219, 48, 122}
\definecolor{mypink3}{cmyk}{0, 0.7808, 0.4429, 0.1412}
\definecolor{red}{RGB}{255, 0, 0}

\definecolor{blackk}{RGB}{0,0,0}
\definecolor{orangee}{RGB}{230,159,0}
\definecolor{skybluee}{RGB}{86, 180, 233}
\definecolor{bluishgreenn}{RGB}{0,158,115}
\definecolor{yelloww}{RGB}{240,228,66}
\definecolor{bluee}{RGB}{0,114,178}
\definecolor{vermillionn}{RGB}{213,94,0}
\definecolor{reddishpurplee}{RGB}{204,121,167}

\colourcheck{bluishgreenn}

\title{\LARGE \bf
A Resource Efficient Fusion Network for Object Detection in \\Bird's-Eye View using Camera and Raw Radar Data}

\author{Kavin Chandrasekaran$^{1,2\,*}$, Sorin Grigorescu$^{1,3}$, Gijs Dubbelman$^{2}$, Pavol Jancura$^{2}$
\thanks{$^{*}$co-first author}
\thanks{$^{1}$Elektrobit Automotive GmbH 
            {\tt\small{\{kavin.chandrasekaran, sorin.grigorescu\}@elektrobit.com}}}
\thanks{$^{2}$Eindhoven University of Technology
            {\tt\small{\{k.chandrasekaran, g.dubbelman, p.jancura\}@tue.nl}}}
\thanks{$^{3}$Transilvania University of Brasov
            {\tt\small{\{s.grigorescu\}@unitbv.ro}}}
}

\begin{document}

\maketitle
\thispagestyle{empty}
\pagestyle{empty}

\begin{abstract}
Cameras can be used to perceive the environment around the vehicle, while affordable radar sensors are popular in autonomous driving systems as they can withstand adverse weather conditions unlike cameras. However, radar point clouds are sparser with low azimuth and elevation resolution that lack semantic and structural information of the scenes, resulting in generally lower radar detection performance. In this work, we directly use the raw range-Doppler (RD) spectrum of radar data, thus avoiding radar signal processing. We independently process camera images within the proposed comprehensive image processing pipeline. Specifically, first, we transform the camera images to Bird's-Eye View (BEV) Polar domain and extract the corresponding features with our camera encoder-decoder architecture. The resultant feature maps are fused with Range-Azimuth (RA) features, recovered from the RD spectrum input from the radar decoder to perform object detection. We evaluate our fusion strategy with other existing methods not only in terms of accuracy but also on computational complexity metrics on RADIal dataset.

\end{abstract}

\section{INTRODUCTION}
\label{introductionsec}
Autonomous Driving Systems (ADS) often rely on different types of sensors to achieve accurate perception. Most of the self-driving vehicles are equipped with cameras, radars, and LiDARs~\cite{yeong_sensor_2021}. Camera data provide rich visual information about the environment. However, they are susceptible to bad weather conditions and lack depth perception capabilities. On the contrary, the expensive LiDAR sensor provides denser point clouds with precise depth and spatial information, delivering a higher resolution detail for objects in 3D space compared to camera images or sparser radar points. Both the camera and the LiDAR suffer in adverse weather conditions like fog, smog, and snowstorms~\cite{avidan_radatron_2022}. 

On the other hand, the radar point cloud data processed typically from constant false alarm rate (CFAR~\cite{richards_principles_2010}) algorithm suffers low angular resolution and severe sparsity as contextual information of radar returns are lost~\cite{lim_radar_2019}. In contrast, leveraging raw radar data is considered to hold significant potential for perception tasks. Thankfully, radar and camera technologies complement each other significantly, making their fusion a promising solution to common perception tasks, specifically object detection. Nevertheless, effectively employing the raw radar data, particularly for fusion with other sensors, remains a challenge.

\begin{figure}
    \centering 
    \includegraphics[width=8.5cm, height=5cm]{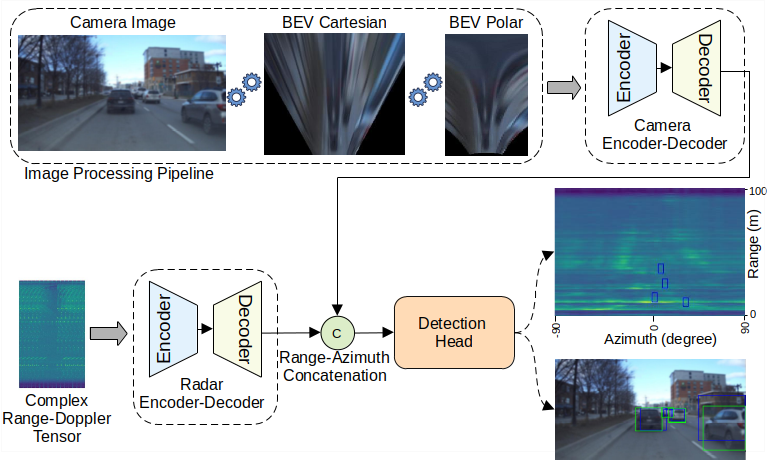}
    \caption{\textbf{Architecture Overview:} The image processing pipeline first transforms the camera image into Bird's-Eye View (BEV). Subsequently, the resultant BEV undergoes conversion into polar representation, directly mapping to the Range-Azimuth (RA) image. Object detection is performed on RA image features fused with radar features from the radar decoder. The predictions obtained in the RA view are shown in the camera images with ground-truth bounding boxes in green and predictions in blue.}
    \label{figarch:introoverview}
\end{figure}

Recently, the success of Bird's-Eye View (BEV) based methods have sparked a surge of research interest, particularly for its effectiveness in perception tasks. Using transformations to project 2D image features into 3D BEV space is one approach as proposed in LSS~\cite{philion_lift_2020} and OFT~\cite{roddick_orthographic_2018}. Another approach involves the utilization of initialized BEV queries~\cite{zhou_cross-view_2022} or object queries~\cite{liu_petr_2022, liu_petrv2_2022} to iteratively extract features from multiview images. Drawing on these sophisticated methodologies, BEVFusion~\cite{liu_bevfusion_2023} delves into the benefits of utilizing BEV representation in multisensor fusion, resulting in remarkable performance outcomes. BEVFormer~\cite{li_bevformer_2022} and its variant~\cite{yang_bevformer_2023} leverage temporal information to improve detection capability, whereas BEVStereo~\cite{li_bevstereo_2022} and STS~\cite{wang_sts_2022} investigate the advantages of estimated depth for BEV-based detection. The recent UniTR~\cite{wang_unitr_2023} employs modality-agnostic transformer encoder to manage view-discrepancy in sensor data. Nevertheless, researchers are also intrigued by how to effectively utilize other sources of sensor information to their fullest potential. 

In this work, we explore the potential of raw radar data from the latest RADIal dataset~\cite{rebut_raw_2022} by adopting their High Definition (HD) radar sensing model, FFTRadNet and extend their network for fusion with camera images in the BEV polar representation as shown in Fig.~\ref{figarch:introoverview}. 

The contributions of our work can be summarized as follows: \textbf{(I)} We propose a new fusion architecture setup to effectively learn the BEV image and radar features in the \textit{polar domain}. \textbf{(II)} This is achieved by first processing the front-view camera images to Range-Azimuth (RA) like representation in a separate image processing pipeline. \textbf{(III)} Extensive analysis shows that our \textit{compute resource efficient} method closely competes with other state-of-the-art models in accuracy and surpasses them in computational complexity. The code is made publicly available through \href{https://github.com/tue-mps/refnet}{https://github.com/tue-mps/refnet}.

\section{RELATED WORK} \label{relatedwork}

\subsection{Camera-Radar dataset} \label{relatedwork_dataset}

To enable effective sensor fusion, it is essential that the sensor data streams are synchronized both temporally and spatially. Additionally, precise calibration parameters for each sensor involved in the fusion process must be accurately determined and known. Recently, there has been increasing attention to leveraging radar data beyond point cloud representations for fusion with camera images for enhancing object detection. As a result, some attempts have been made in providing radar data in the form of range-azimuth-Doppler (RAD) tensor~\cite{ouaknine_carrada_2021, zhang_raddet_2021}, range-azimuth (RA) maps~\cite{sheeny_radiate_2021}, range-Doppler (RD) spectrum~\cite{mostajabi_high_2020} or even raw Analog Digital Converter (ADC) data~\cite{mostajabi_high_2020, lim_radical_2021}. All the above forms can be derived from ADC data using Fast Fourier Transform (FFT).

Our motivation is to leverage the potential of raw radar data due to its comprehensive representation instead of sparser point clouds. Recent datasets that offer such a representation are RADIal~\cite{rebut_raw_2022}, Radatron~\cite{avidan_radatron_2022}, RADDet~\cite{zhang_raddet_2021}, K-Radar~\cite{paek_k-radar_2023}. The RADIal benchmark~\cite{rebut_raw_2022} has been chosen for this research, as it is the only dataset providing an analog-to-digital converter (ADC) signal, Range-Angle-Doppler (RAD) tensor, Range-Angle (RA) view, Range-Doppler (RD) view, point cloud (PC) representation of HD radar data, combined with camera, LiDAR, and odometry. This implies that there are greater opportunities to explore various fusion settings. However, the scope of detection is limited to vehicle class, since the majority of road users in this dataset are several moving vehicles. But it is also possible to extend our work to other road users given a suitable dataset.

\subsection{Camera-Radar fusion methods for object detection} \label{relatedwork_crmethods}

Methods utilizing radar point cloud face challenges due to sparsity and low angular resolution. On the other hand, storage and computation pose a concern when using RAD tensors. As a result, RADDet~\cite{zhang_raddet_2021} takes the Doppler dimension as channels and~\cite{major_vehicle_2019} utilizes RAD tensors and projects it to multiple 2D views. RAD tensors are divided into small cubes in RTCNet~\cite{palffy_cnn_2020} and to reduce computational efforts, 3D CNNs are applied. Apart from that, the networks of~\cite{rebut_raw_2022, zhang_object_2020} intakes complex RD spectrum to extract spatial information. But RODNet~\cite{wang_rodnet_2021} operates using RA maps for detection, preventing false alarms caused by extended Doppler profile. Recently, utilization of ADC data has gained attention~\cite{giroux_t-fftradnet_2023, yang_adcnet_2023} with minimal success. 

The aforementioned methods operate on raw radar data as standalone inputs. Fusing various sensor data yields complementary cues, thereby enhancing performance robustness. The approaches of~\cite{chadwick_distant_2019, nobis_deep_2020} fuses camera images and projected radar point cloud data in a perspective space at input level. ~\cite{nabati_radar-camera_2020, nabati_centerfusion_2021} target to fuse at the Region of Interest (RoI) level, while~\cite{kim_grif_2020, kim_crn_2023} combine RoIs generated independently by different sensors. Architectures like~\cite{zhang_rvdet_2021, wu_mvfusion_2023} perform feature level fusion by integrating the feature maps from different modalities, while~\cite{kim_low-level_2020, kim_craft_2022} use RoIs to crop and unify features across modalities.

The recent architectures that use RADIal~\cite{rebut_raw_2022} dataset to perform fusion are Cross Modal Supervision (CMS)~\cite{jin_cross-modal_2023}, ROFusion~\cite{liu_rofusion_2023} and EchoFusion~\cite{liu_echoes_2023}. CMS~\cite{jin_cross-modal_2023} uses the RD spectrum and takes support from pseudo-labels generated from camera images, while ROFusion~\cite{liu_rofusion_2023} associates the RD and image feature maps by additionally using radar points that are located within the bounding box labels, proposing a hybrid point-wise fusion strategy.  EchoFusion~\cite{liu_echoes_2023} on the other hand, uses Polar Aligned Attention technique that fuses the features from Range-Time (RT) radar maps and images in a unified Bird's-Eye View (BEV) perspective, which also demonstrated the potential of radar as a low-cost alternative to LiDAR. 

In contrast to the aforementioned methods, we preprocess the camera data prior to feeding them into our network. Our image processing technique involves the transformation of the front-view camera image to a BEV radar-like representation in the polar domain as described in Section~\ref{imageprocessing}, and the detailed fusion setup is described in Section~\ref{polarsegfusionnetarchi} with experimental setup in Section~\ref{experiments}.

When employing raw radar data, it is important to take into account computational complexity metrics. However, previous studies often lack detailed reporting on the resource demands of their models. In this work, we not only compare the performance of our approach with other models in terms of accuracy, but also on resource consumption, as pointed out in Section~\ref{results}.

\section{METHOD} \label{method}

\subsection{Problem statement} \label{problemstatement}
The problem statement revolves around the need to present an architecture that solves the proposed perception task resource efficiently by introducing an independent image processing pipeline. 

\begin{figure*}[!ht]
    \centering 
    \includegraphics[width=\textwidth]{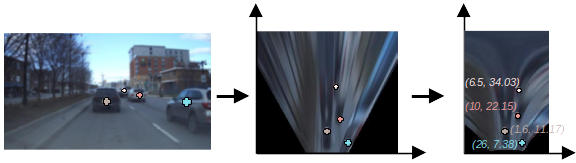}
    \put(-450,120){\fontsize{12}{12}\selectfont \textbf{Camera Image}}
    \put(-260,120){\fontsize{12}{12}\selectfont \textbf{BEV Cartesian}}
    \put(-90,120){\fontsize{12}{12}\selectfont \textbf{BEV Polar}}
    \put(-170,-2){\fontsize{12}{12}\selectfont \textit{x (m)}}
    \put(-310,120){\fontsize{12}{12}\selectfont \textit{y (m)}}
    \put(-35,-2){\fontsize{12}{12}\selectfont \textit{$\theta$ (deg)}}
    \put(-130,120){\fontsize{12}{12}\selectfont \textit{r (m)}}
    \put(-315,68){\fontsize{10}{10}\selectfont \textit{\textbf{Step 1}}}
    \put(-145,68){\fontsize{10}{10}\selectfont \textit{\textbf{Step 2}}}
    \put(-430,0){\fontsize{10}{10}\selectfont 540x960x3}
    \put(-239,-1){\fontsize{10}{10}\selectfont 216x250x3}
    \put(-85,-1){\fontsize{10}{10}\selectfont 512x256x3}
    
    \caption{\textbf{Image Processing Pipeline:} The objects in the frame \textit{(four cars)} marked in different colors are reflected in the BEV Cartesian and Polar pixel images. The origin is at the bottom center. The azimuth $(\theta)$, range $(r)$ ground truth polar coordinates are marked for reference. $r$ denotes the distance from the objects to the ego vehicle (in meters); $\theta$ represents the angle at which the objects are located in degrees.}
    \label{figarch:imagepreprocessing}
\end{figure*}
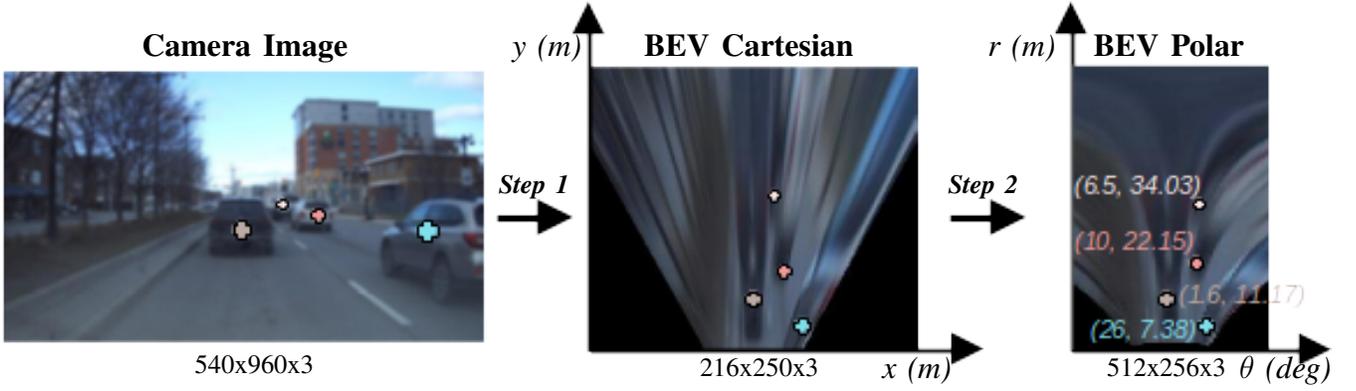

\subsection{Image processing pipeline} \label{imageprocessing}
The camera images are typically recorded in perspective view, while radar data can be transformed from raw ADC signal to Range-Angle-Doppler (RAD) tensor, Range-Angle (RA) view, Range-Doppler (RD) view, or Point Cloud (PC) representation. Thus, it is crucial to identify a shared representation such that the sensor data fusion can be achieved to perform the intended task. 

Processing High Definition (HD) radar data, renowned for its enhanced resolution and complexity, places a considerable demand on computational resources. Consequently, in this approach, we transform the camera image to an RA like representation that entails less intensive computational requirements. This transformation involves two steps, as shown in Fig.~\ref{figarch:imagepreprocessing}. We emphasize that a taxonomy of algorithms are presented in a recent survey~\cite{ma_vision-centric_2023} that includes our inspiration PolarFormer~\cite{jiang_polarformer_2023} which performs object detection in BEV Polar coordinate.

\textit{\textbf{Step 1:}} We are given a set of $N$ camera input images with training sample index $n$, \{$I_{img}^{n} \in \mathbb{R}^{H \times W \times 3}$\}$_{n=1}^{N}$. As a first step, the camera image ($I_{img}^{n}$) is converted to BEV Cartesian domain image~\cite{noauthor_transform_nodate}. Given the camera intrinsic parameters $(\Pi \in \mathbb{R}^{3 \times 3})$, height at which the camera is mounted $(h)$ and pitch value $(p)$, the image formation is, $\varepsilon$. 
\begin{align}
    \varepsilon = f(\Pi,h,p)
\end{align}

Given $\varepsilon$, a BEV Cartesian object is created which is used to transform the given front-view camera image:

\begin{align}
    BEV_{Cartesian} = f_{Image}^{Cartesian}(\varepsilon, \eta, O, I_{img}^{n})
\end{align} where, $\eta = [5, 50, -22, 22]$ corresponds to the output space as [xmin, xmax, ymin, ymax] in vehicle coordinate system. The X-axis oriented forward from the vehicle and Y-axis oriented towards the left. $O \in \mathbb{R}^{rows \times cols}$ is specified as output image size [nrows, ncols] in pixels. 

\textit{\textbf{Step 2:}} The obtained BEV Cartesian image is further converted to the Polar domain \(f_{Cartesian}^{Polar}[BEV_{Cartesian}]\). Firstly, all Cartesian pixel indices $(x, y)$ are transformed to Polar indices ($\theta, r$) using the following equations:

\begin{align}
    \theta &= \arctan2(y, x) \\
    r &= \sqrt{x^2 + y^2}
\end{align}

The obtained Polar indices ($\theta, r$) are projected back to pixel coordinates ($\theta_{pixel}, r_{pixel}$) for an an image like representation as follows:
\begin{align}
    \theta_{pixel} &= r \cdot \sin(\theta) \\
    r_{pixel} &= r \cdot \cos(\theta) 
\end{align}

Using these pixel coordinate values, each channel of the camera image array is projected individually to the BEV Polar image using spline interpolation~\cite{noauthor_scipyndimagemap_coordinates_nodate} and restacked. This approach consumes less memory than projecting the three-dimensional image array in one step.

This whole independent process of pre-aligning camera images in BEV Polar space offers multiple advantages. Mainly, it eliminates the necessity for feature transformation within the network, potentially enhancing computational efficiency during the training process.

\subsection{Architecture design} \label{polarsegfusionnetarchi}

\begin{figure*}[!ht]
    \centering 
    \includegraphics[width=\textwidth]{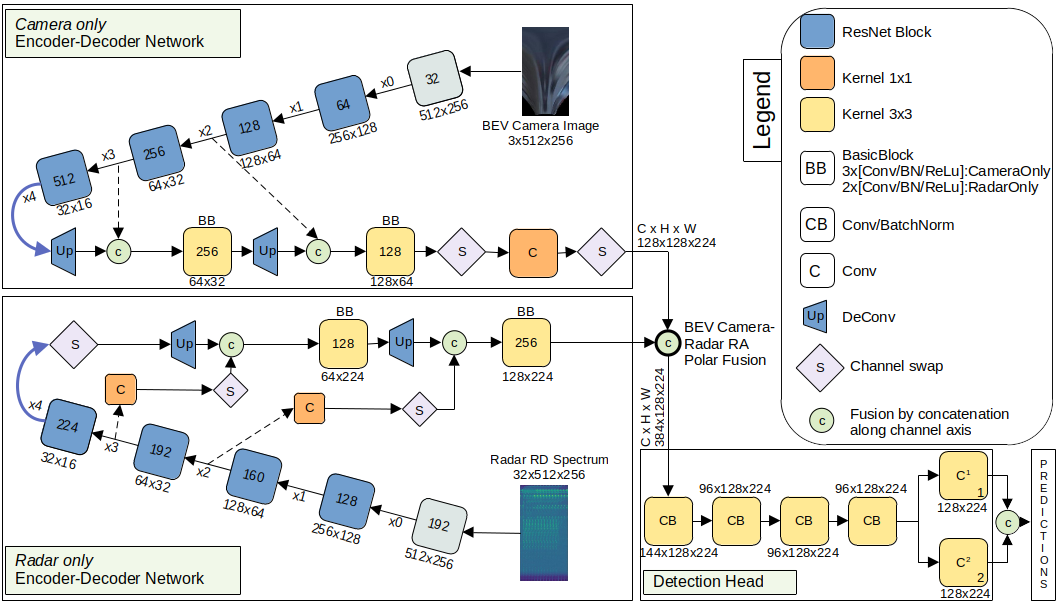}
    \caption{The \textit{camera only} and \textit{radar only} encoder contains four ResNet-50-like blocks with a pre-encoder block. The features from each of those blocks are named x0, x1, x2, x3, and x4. The thick blue curved arrow takes the encoder's output to the decoder's input in order to expand the input feature maps to higher resolutions. The dotted lines represent the skip connections used to preserve spatial information. The features from the \textit{camera only} decoder and \textit{radar only} decoder are then \textit{fused} before passing them to the detection head. The head finally predicts the objects in Bird's Eye RA Polar View, as shown in Fig.~\ref{figarch:introoverview}.}
    \label{figarch:afterdecoderfusion}
\end{figure*}

Fig.~\ref{figarch:afterdecoderfusion} shows the architecture with radar \textit{(bottom block)} and camera \textit{(top block)} network. 

\textbf{\textit{Radar feature extractor:}} The computational cost involved in processing the raw range-azimuth Doppler (RAD) 3D tensor is higher. Therefore, the use of a denser Range Doppler (RD) map is an alternative consideration, especially when there is a possibility to recover the angle information from RD maps~\cite{rebut_raw_2022}.

The radar sensor used comprises 12 transmitting antennas and 16 receiving antennas, resulting in a total of 16 channels within the input tensor. This means the signature of any object, say a vehicle in front, will be visible 12 times for each receiving antenna. In particular, it will be measured at range-Doppler positions $(R, (D+k\Delta)[D_{\text{max}}])_{k=1,\ldots,12}$ , where $\Delta$ denotes the Doppler shift that is caused by the phase shift $\Delta_\varphi$ in the transmitted signal. $D_{max}$ is the maximum measurable Doppler value. If the measured Doppler value $(D+k\Delta)$ exceeds $D_{\text{max}}$, it will be truncated to fit within $D_{\text{max}}$, ensuring that all measured Doppler values fall within this allowable range. 

The Range-Doppler (RD) tensor hence is organized as complex numbers $(R+iD)$ representing Range and Doppler values. Rearranging and concatenating the real and imaginary parts of this tensor results in 32 channels for the input with 512 and 256 range and Doppler bins, respectively. To this end, we adapt a Multiple Input Multiple Output (MIMO) pre-encoder~\cite{donnet_mimo_2006} that reorganizes this RD tensor into a meaningful representation for the resnet-50~\cite{he_deep_2016} like encoder blocks with 3, 6, 6 and 3 residual layers respectively from FFTRadNet~\cite{rebut_raw_2022}.

More specifically, the extracted feature tensors can be viewed as azimuth, range, Doppler, respectively. Since the objective is to acquire the angle information, \textit{channel swapping strategy} is employed by swapping the Doppler and azimuth axes before upscaling the feature maps. This is depicted with a rhombus highlighted in purple in Fig.~\ref{figarch:afterdecoderfusion}. As a result, we seek to learn a dense feature embedding of RA maps, thus recognizing their relevance to the subsequent object detection task.

\textbf{\textit{Camera feature extractor:}} As explained in Section~\ref{imageprocessing}, the camera images are processed (refer Fig.~\ref{figarch:imagepreprocessing}) to obtain a Bird's-Eye View RA Polar representation. This representation is the input to our \textit{camera only} CNN model. We have chosen this representation as it directly relates to the decoded features of the \textit{radar only model}, which in turn supplements the radar features upon fusion, as shown by a thick black fusion circle in Fig.~\ref{figarch:afterdecoderfusion}.

The \textit{camera only} model starts with a pre-encoder block that performs an initial feature extraction with a standard kernel size of 3. Our Feature Pyramid Network (FPN) encoder is composed of 4 blocks with 3, 6, 6, and 3 residual layers, respectively~\cite{he_deep_2016}. Each encoder block here performs a ${2{\times} 2}$ downsampling which leads to a reduction of tensor size by a factor of 16 in height and width. This downsampling is to prevent losing the signature of small objects that are few pixels in BEV Polar image.

Channel swapping strategy is not required within the decoder of \textit{camera only} network, since the extracted features already take the form of Range-Azimuth (RA) like representation. It is important to understand that the channels are still swapped twice for a different purpose. After the second basic block (BB in Fig.~\ref{figarch:afterdecoderfusion} represents Basic Block), the channels are swapped ${(128{\times} 128{\times} 64 \stackrel{\text{swap}}{\longrightarrow} 64{\times} 128{\times} 128)}$ to increase dimension of the azimuth axis ${(64{\times} 128{\times} 128 \stackrel{\text{conv2d}}{\longrightarrow} 224{\times} 128{\times} 128)}$ using a convolutional layer. Further, it is again swapped back so that the RA tensor is regained. This strategy allows to view the decoded feature map in a dimension ${(128{\times} 128{\times} 224)}$ that helps the network in effectively fusing with the radar features during training by reducing computational overhead. Please note that our \textit{camera only} architecture backbone module could be replaced by heavier models depending on computational budgets (refer to Section~\ref{ablationstudy}).

\textbf{\textit{Fused inputs to the detection head:}} The RA latent features from the \textit{camera only} and \textit{radar only} network are fused by channel concatenation, which is then processed using four Conv-BatchNorm layers having 144, 96, 96 and 96 filters respectively. The branch bifurcates into classification and regression pathways. The classification segment (C$^{1}$) consists of a ${3{\times} 3}$ convolutional layer with sigmoid activation, predicting a probability map. Each pixel in this map corresponds to binary classification as occupied by a vehicle or not. The dimension of the predictions from classification pathway is ${1{\times} 128{\times} 224}$ with a resolution of 0.8m in range and 0.8$^{\circ}$ in azimuth. This low resolution in dimension is to reduce the computational complexity and is enough to distinguish two objects. The regression part (C$^{2}$) employs the same ${3{\times} 3}$ convolution layer that outputs two feature maps which predict the final range and azimuth values of the detected objects. Hence, this takes the shape ${2{\times} 128{\times} 224}$, where the channels correspond to range and azimuth values.

\section{EXPERIMENTAL SETUP} \label{experiments}

\subsection{Dataset} \label{radialdataset}

The RADIal dataset is a compilation of two hours of raw data from synchronized automotive-grade sensors (camera, laser, and High Definition radar) in a variety of settings (city street, highway, country road) that includes GPS data. The three sensors are synchronized for about 25,000 frames, of which 8,252 are labeled with a total of 9,550 vehicles. Recent radar-camera fusion surveys~\cite{yao_radar-camera_2023, hu_survey_2020, wang_multi-sensor_2020} compare the other relevant publicly available dataset.

\subsection{Training details} \label{traindetails}

The network was trained on a workstation equipped with an Intel Core i9-10940X CPU, a Nvidia RTX A6000 GPU and 52 GB RAM. The dataset is randomly split so that 70\% corresponds to the training data and the remaining is the validation and test set (approx. 15\% each). Training is carried out for 100 epochs using Adam optimizer~\cite{kingma_adam_2017} with mini-batches of size 4. The initial learning rate of 1e-4 has been set with a decay of 0.9 for every 10 epochs. 

Since a large proportion of the scene belongs to background, we use focal loss~\cite{lin_focal_2018} on the classification output, so that the training process is stabilized thus avoiding the class imbalance issue. The smooth L1 loss is used on the regression output specifically for positive detections. 
\begin{equation}
\mathcal{L}_{det} = \text{Focal}(y_{\text{cls}}, \hat{y}_{\text{cls}}) + \alpha \text{Smooth-L1}(y_{\text{reg}}, \hat{y}_{\text{reg}})
\end{equation} where $y_{\text{cls}}$ and $\hat{y}_{\text{cls}}$ represent the ground truth and predicted values for classification, respectively. $y_{\text{reg}}$ and $\hat{y}_{\text{reg}}$ represent the ground truth and the predicted values for regression part, respectively. $\alpha$ is a hyperparameter that balances the contributions of the two loss components. Please refer~\cite{yang_pixor_2019} for details.

\subsection{Evaluation metric} \label{evalmetrics}

We use Average Precision (AP) and Average Recall (AR) with an Intersection-Over-Union (IoU) threshold of 50\% as our accuracy metrics in all experiments. F1-Score is computed from AP and AR directly: \( F1 = \frac{2 \times \text{AP} \times \text{AR}}{\text{AP} + \text{AR}} \). We also present the absolute Range (in meters) and Angle Error (in degrees) as follows:

\begin{align}
RE = \frac{\sum\limits_{n=0}^{N} \sum\limits_{m=0}^{M} (|r_m - \hat{r}_m|)}{m_{total}}; AE = \frac{\sum\limits_{n=0}^{N} \sum\limits_{m=0}^{M} (|\theta_m - \hat{\theta}_m|)}{m_{total}}
\end{align} where, $r_m$, $\hat{r}_m$ are the ground truth and the predicted range values of an object in meters. $\theta_m$, $\hat{\theta}_m$ are the ground truth and the predicted azimuth values of an object in degrees. M denotes the number of objects in a particular frame. N denotes the total number of frames in the test data. $m_{total}$ is the total number of objects in the test data. The AP, AR, Range Error (RE) and Angle Error (AE) are computed for different classification confidence score thresholds from 0.1 to 0.9 with a step of 0.1 and averaged by following the official implementation~\cite{rebut_raw_2022}.

Additionally, we evaluate the model's complexity by comparing the total number of trainable parameters (\#: in millions), the Average Frames Per Second (FPS); for a given model, an FPS value is computed for each frame in the test set and averaged. We also calculate the standard deviation ($\sigma$) from the FPS values, where a lower $\sigma$ indicates a more consistent performance across frames. Furthermore, we consider the size of the model in megabytes (MB) and GPU memory cost in gigabytes (GB) as shown in Table~\ref{table:latefusion}.

\subsection{Baselines}
\label{baselines}
Most of the methods presented in Sections~\ref{introductionsec} and~\ref{relatedwork} rely on low-resolution traditional radars and can encounter difficulty in accommodating HD radar data due to memory constraints. As HD radar is used in this work, we consider as baselines the state-of-the-art models that have been trained on the RADIal dataset. As discussed in Section~\ref{relatedwork}, FFTRadNet~\cite{rebut_raw_2022}, TFFTRadNet~\cite{giroux_t-fftradnet_2023}, ADCNet~\cite{yang_adcnet_2023} are some of the works that fall under this category. Since we focus on camera and raw radar data fusion in Bird's-Eye View for object detection, the Cross-Modal Supervision (CMS)~\cite{jin_cross-modal_2023}, ROFusion~\cite{liu_rofusion_2023} and EchoFusion~\cite{liu_echoes_2023} are closely related.

\section{RESULTS} \label{results}
This section conducts a comprehensive evaluation of our model, featuring quantitative and qualitative results analysis with visuals of predictions.

\subsection{Quantitative Evaluation}
We compare our model performance not only on accuracy metrics, but also on the computational parameters presented in Section~\ref{evalmetrics}. The GPU cost is influenced by various hyperparameters, with batch size being one of the most significant factors. We used the same training parameters as other models for a fair comparison as presented in Section~\ref{traindetails}.

\textit{\textbf{Accuracy:}} We outperform the existing fusion detection frameworks in Range and Angle Error which indicates that the detected objects are accurately localized in the scene. Achieving second best F1-Score and marginal difference in AP and AR could be attributed to the more complex architecture of EchoFusion~\cite{liu_echoes_2023}, which has nearly four times the number of trainable parameters compared to our model.

The large improvement in Angle Error (AE) could be due to the fact that the frames are \textit{preprocessed} in an independent image processing pipeline \textit{(refer Section~\ref{imageprocessing})}, performing no image feature transformations from Cartesian to Polar during training. Radar's RA view is inherently polar and hence we convert the camera image to polar view.

Despite our efforts to train the model using Cartesian domain camera data, the results were suboptimal as shown in Table~\ref{table:cartesian}.

\begin{table*}[!ht]
\caption{\label{table:latefusion} Vehicle detection performances on the RADIal dataset test split. RD, ADC, RPC, RT, C correspond to Range-Doppler, Analog-To-Digital Converter signal, Radar Point Cloud, Range-Time signal, and Camera data respectively. Best values are in bold and second best are underlined. \dag: reimplemented with only detection head as they are multi-tasking models. The missing values are indicated by a "-", either due to code unavailability or unreported in the respective works.}
\centering 
\renewcommand{\arraystretch}{1}
\begin{tabular}{llllllllllll}
\specialrule{0.5pt}{0pt}{0pt}
\hline
\textbf{} & \textbf{} & \multicolumn{5}{c}{\multirow{1.25}{*}{\centering{Accuracy Metrics}}} & \multicolumn{5}{c}{\multirow{1.25}{*}{\centering{Computational Metrics}}}
\\\cmidrule(lr){3-7} \cmidrule(lr){8-12}
Methods & Modality & AP(\%) $\uparrow$ & AR(\%) $\uparrow$ & F1(\%) $\uparrow$ & RE(m)$\downarrow$ & AE($^{\circ}$)$\downarrow$ & \#$\downarrow$ & \begin{tabular}[c]{@{}c@{}}Avg \\ FPS\end{tabular} $\uparrow$ & $\sigma$ $\downarrow$ & \begin{tabular}[c]{@{}c@{}}Model\\size\end{tabular}$\downarrow$ & \begin{tabular}[c]{@{}c@{}}GPU \\cost\end{tabular}$\downarrow$ \\\hline \addlinespace[3pt]
FFTRadNet\dag~\cite{rebut_raw_2022}             & RD                     & 93.45        & 83.35        & 88.11             & \underline{0.12}            & 0.15              & \textbf{3.23}        & \textbf{68.46}               & 2.19         & \textbf{39.2}          & \textbf{2.01}                                                    \\
TFFTRadNet\dag~\cite{giroux_t-fftradnet_2023}              & ADC                             & 90.80         & 88.31         & 89.54              & 0.15            & 0.13              & 9.08        & 54.37               & {4.28}         & 109.5         & \underline{2.04}                                                         \\
ADCNet~\cite{yang_adcnet_2023}            & ADC                                 & 95           & 89           & 91.9              & 0.13            & \underline{0.11}              & -           & -                & -            & -             & -                                                            \\
CMS~\cite{jin_cross-modal_2023}            & RD\&C                      & \underline{96.9}         & 83.49        & 89.69             & 0.45            & -                 & 7.7         & -               & -            & -             & -                                                            \\
ROFusion~\cite{liu_rofusion_2023}      & RD\&RPC\&C                       & 91.13        & \textbf{95.29}        &{93.16}             & 0.13            & 0.21              & \underline{3.33}        & 56.11               &\underline{1.55}         & 87.2          & 2.87                                                         \\
EchoFusion~\cite{liu_echoes_2023}      & RT\&C                            & \textbf{96.95}        & \underline{93.43}        & \textbf{95.16}             & \underline{0.12}            & 0.18              & 25.61       & -                & -            & 102.5         & -                                                            \\\hline \addlinespace[3pt]

Ours      & RD\&C                              & 95.75        & {91.35}        & \underline{93.49}             & \textbf{0.11}          & \textbf{0.09}              & 6.58        & \underline{58.91}               & \textbf{1.28}         & \underline{79.8}          & {2.06}          
                                                        
\\\hline    
\specialrule{0.5pt}{0pt}{0pt}
\end{tabular}
\end{table*}

\begin{table}[!ht]
\caption{\label{table:cartesian} BEV Cartesian vs. Polar image input.}
\centering 
\renewcommand{\arraystretch}{1}
\begin{tabular}{llllll}
\specialrule{0.5pt}{0pt}{0pt}
\hline \addlinespace[2pt]
Methods & AP(\%) $\uparrow$ & AR(\%) $\uparrow$ & F1(\%) $\uparrow$ & RE(m)$\downarrow$ & AE($^{\circ}$)$\downarrow$ \\\hline \addlinespace[2pt]
Ours$_{Cart}$      & 90.72  & 86.21     & 88.41             & 0.12        & 0.11
\\\hline \addlinespace[2pt]
Ours$_{Polar}$      & \textbf{95.75}  & \textbf{91.35}    & \textbf{93.49}             & \textbf{0.11}        & \textbf{0.09}    
\\\hline    
\specialrule{0.5pt}{0pt}{0pt}
\end{tabular}
\end{table}
\begin{figure*}[!ht]
    \centering 
    \vspace{0.18cm} 
\begin{subfigure}{0.25\textwidth}
  \includegraphics[width=2.3cm, height=4.3cm, angle=90]{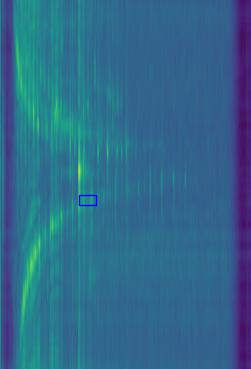}
\end{subfigure}\hfil
\begin{subfigure}{0.25\textwidth}
  \includegraphics[width=2.3cm, height=4.3cm, angle=90]{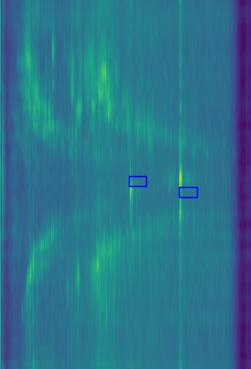}
\end{subfigure}\hfil 
\begin{subfigure}{0.25\textwidth}
  \includegraphics[width=2.3cm, height=4.3cm, angle=90]{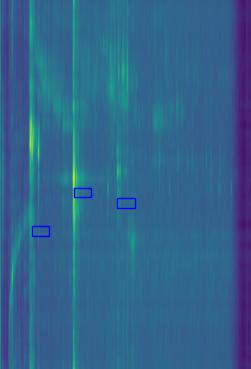}
\end{subfigure}\hfil
\begin{subfigure}{0.25\textwidth}
  \includegraphics[width=2.3cm, height=4.3cm, angle=90]{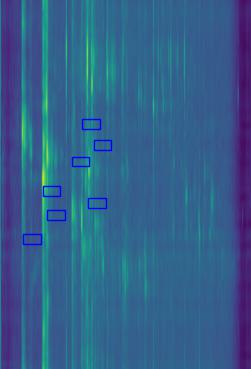}
\end{subfigure}

\medskip
\begin{subfigure}{0.25\textwidth}
  \includegraphics[width=4.3cm, height=2.15cm]{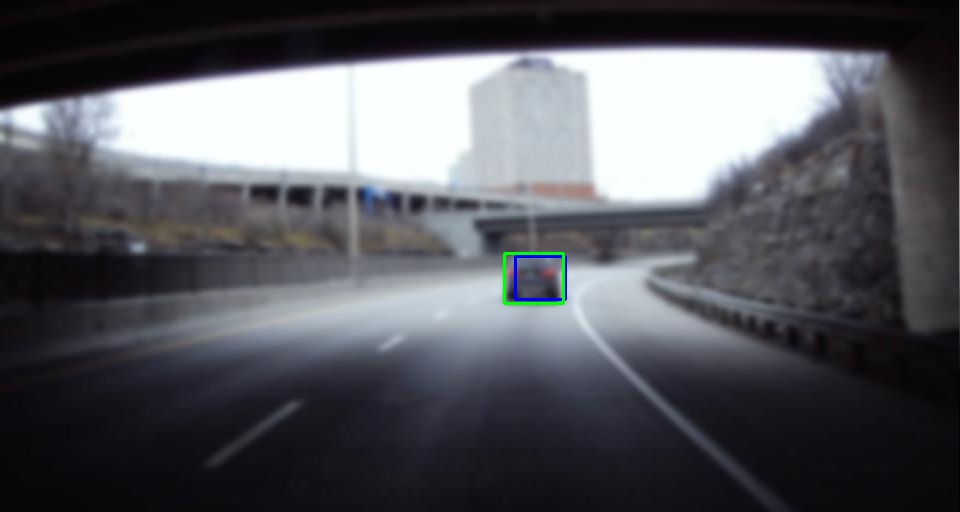}
\end{subfigure}\hfil
\begin{subfigure}{0.25\textwidth}
  \includegraphics[width=4.3cm, height=2.15cm]{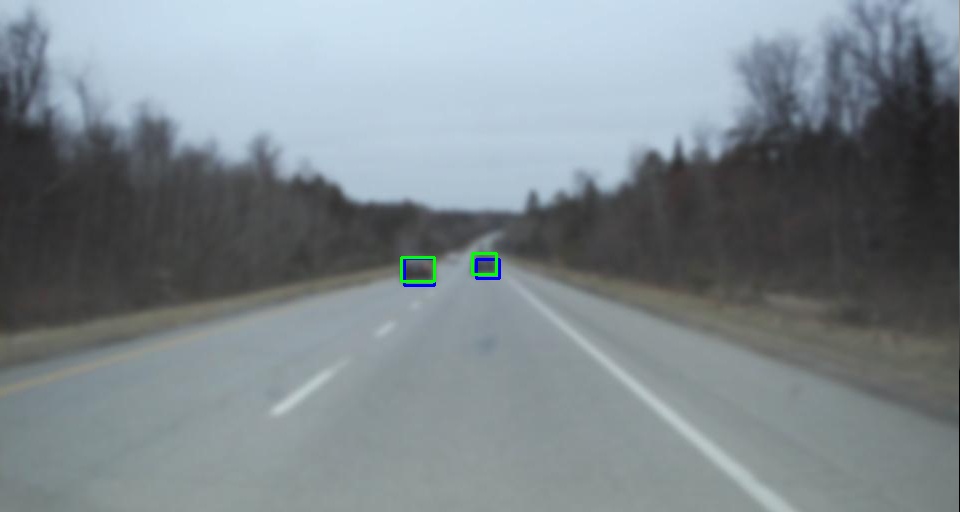}
\end{subfigure}\hfil 
\begin{subfigure}{0.25\textwidth}
  \includegraphics[width=4.3cm, height=2.15cm]{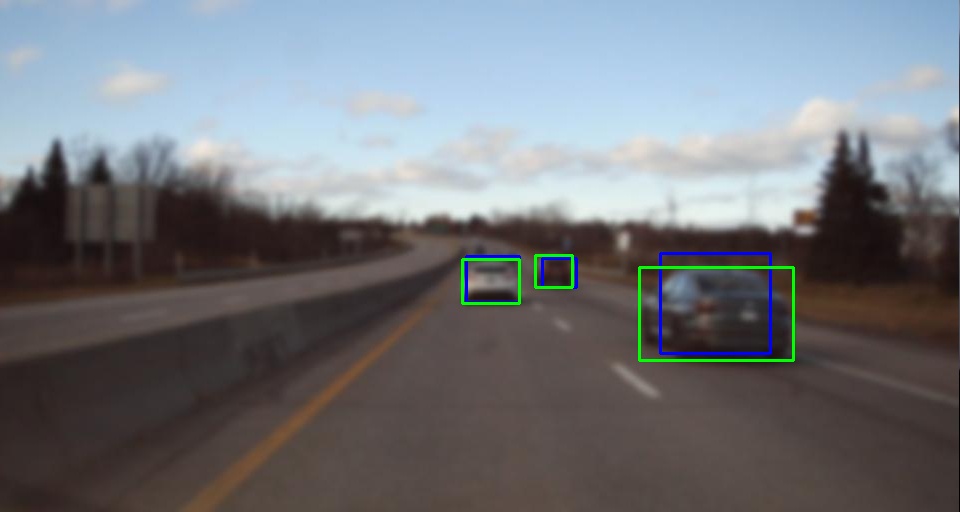}
\end{subfigure}\hfil
\begin{subfigure}{0.25\textwidth}
  \includegraphics[width=4.3cm, height=2.15cm]{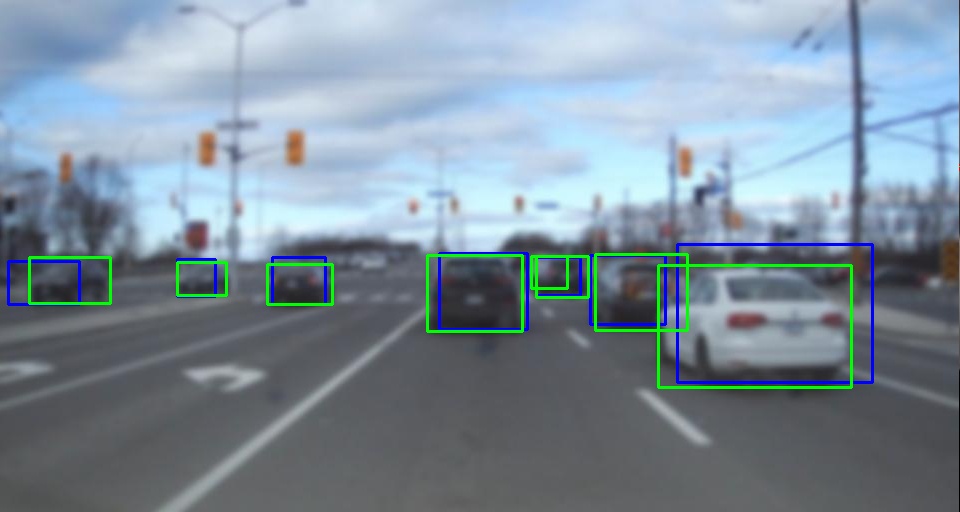}
\end{subfigure}

\caption{Qualitative detection results from the proposed fusion model. The predictions obtained in the RA view (represented as blue boxes in top row) have been projected onto the camera images with ground truth bounding boxes in green.}
\label{fig:qualitativeresults_fusion}
\end{figure*}

\textit{\textbf{Computational complexity:}} FTTRadNet~\cite{rebut_raw_2022} and  TFFTRadNet~\cite{giroux_t-fftradnet_2023} are computationally efficient while compromising on accuracy metrics as they are non fusion models relying only on radar data. On the other hand, models like ROFusion~\cite{liu_rofusion_2023} exhibits inefficiency in GPU memory utilization, while EchoFusion~\cite{liu_echoes_2023} is comparatively very large in size. Alternatively, our fusion model offers enhanced overall performance through the integration of multiple modalities, providing a more comprehensive understanding of the data. This additional benefit outweighs the marginal increase in memory consumption, making our fusion model the superior choice for achieving optimal results. Codes of other algorithms~\cite{yang_adcnet_2023, jin_cross-modal_2023} are not available yet for comparison.

\subsection{Qualitative Evaluation}
Frames with different complexity are chosen, as shown in Fig.~\ref{fig:qualitativeresults_fusion}. These results underscore not only the robustness of the model, but also its effectiveness in providing reliable predictions across different conditions, enhancing its practicality in real-world scenarios.


\section{ABLATION STUDY}
\label{ablationstudy}
To deepen the performance, complexity analysis and validate the choice of our network architecture and its input components, we conduct an ablation study with different backbones, presented in Table~\ref{table:ablation}. Specifically, the ablations are carried out by replacing our camera backbone and not the radar, due to the unique complex alignment of the raw radar data, as explained in Section~\ref{polarsegfusionnetarchi}. We compare the original ResNet-50~\cite{he_deep_2016}, EfficientNet-B2~\cite{tan_efficientnet_2020} and ResNet-18 backbone-based model with a transformer decoder called R18-UNetFormer~\cite{wang_unetformer_2022} with our model results from Table~\ref{table:latefusion}.

It is evident that our lightweight model is resource and storage efficient, performing at a high-speed frame rate of 58.91 FPS.

\begin{table*}[!ht]
\caption{\label{table:ablation} Ablation on different backbone architectures. The best is highlighted in bold. Assessing a model solely based on accuracy metrics is unjust, considering that both EfficientNet-B2 and R18-UNetFormer strike a commendable balance.}

\centering
\renewcommand{\arraystretch}{1}
\begin{tabular}{lllllllllll}
\specialrule{0.5pt}{0pt}{0pt}
\hline
\textbf{} & \multicolumn{5}{c}{\multirow{1.25}{*}{\centering{Accuracy Metrics}}} & \multicolumn{5}{c}{\multirow{1.25}{*}{\centering{Computational Metrics}}}
\\\cmidrule(lr){2-6} \cmidrule(lr){7-11}
Backbone & AP(\%) $\uparrow$ & AR(\%) $\uparrow$ & F1(\%) $\uparrow$ & RE(m)$\downarrow$ & AE($^{\circ}$)$\downarrow$ & \# $\downarrow$ & \begin{tabular}[c]{@{}c@{}}Avg \\ FPS\end{tabular} $\uparrow$ & $\sigma$ $\downarrow$ & \begin{tabular}[c]{@{}c@{}}Model \\ size\end{tabular}$\downarrow$ & \begin{tabular}[c]{@{}c@{}}GPU \\ cost\end{tabular}$\downarrow$ \\\hline \addlinespace[3pt]
ResNet-50~\cite{he_deep_2016}                      & 95.53       & 92.13        & 93.79           & \textbf{0.12}           & \textbf{0.09}             & 53.67      & 18.55               & 4.12        & 645.1          & 2.92   
\\       
EfficientNet-B2~\cite{tan_efficientnet_2020}                      &  \textbf{97.04}       & 90.01        &  93.39           & {0.13}           & 0.10             & \textbf{14.47}      & 51.80               & \textbf{0.94}        & \textbf{159.6}          & \textbf{2.26}   
\\   
R18-UNetFormer~\cite{wang_unetformer_2022}                      & 96.43       & \textbf{92.36}        & \textbf{94.35}            & \textbf{0.12}           & 0.10             & 14.98      & \textbf{57.21}               & 1.08        & 180.6          & 2.31   
\\\hline   
\specialrule{0.5pt}{0pt}{0pt}
\end{tabular}
\end{table*}

\section{LIMITATIONS} \label{limitations}

The height of the camera mounted above the ground and the pitch of the camera toward the ground are important parameters when transforming the front-facing camera data to a BEV object. Furthermore, as stated in Section~\ref{imageprocessing}, the area in front of the camera (0 to 50 meters) as well as to either side of the camera (22 meters) are defined to the best of our knowledge as RADIal camera sensor suite does not have range specification information.

\begin{figure}[!ht]
    \centering
    \hspace{0.2cm}
    \begin{subfigure}{0.25\textwidth}
        \centering
        \includegraphics[width=5.4cm, height=2.7cm]{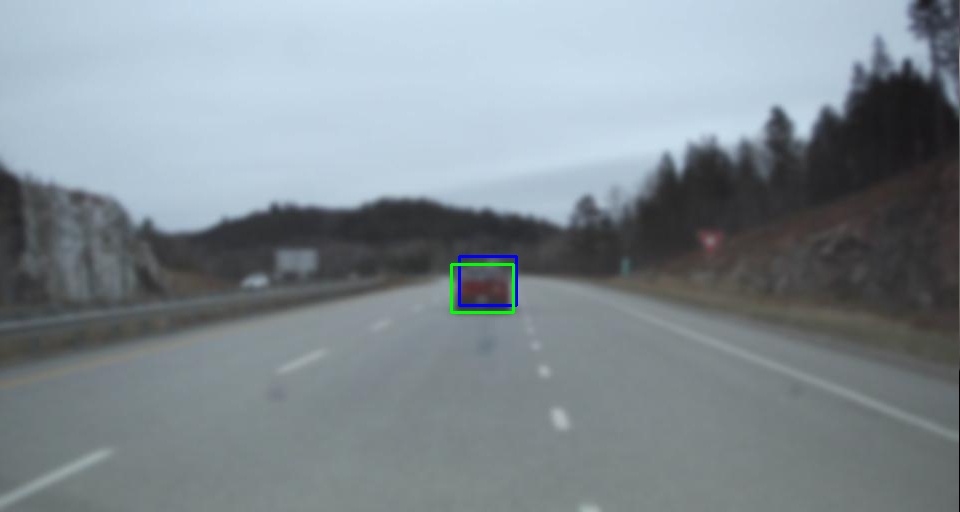}
        \put(-73,-10){\fontsize{10}{10}\selectfont (a)}
    \end{subfigure}
    \hspace{0.5cm}
    \begin{subfigure}{0.15\textwidth}
        \centering
        \includegraphics[width=1.62cm, height=2.7cm]{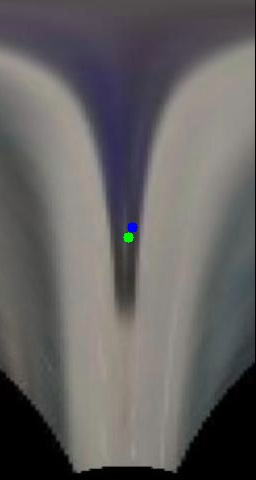}
        \put(-27,-10){\fontsize{10}{10}\selectfont (b)}
    \end{subfigure}
    \caption{\small The prediction in blue and the ground truth in green are shown in (a) front-view camera and (b) BEV Polar image. Zoom in to better visualize.}
    \label{fig:limitations}
\end{figure}

Also theoretically, the green ground truth point in the BEV image in Fig.~\ref{fig:limitations} is affected by an offset. There could be one or several reasons for this discrepancy, such as a transformation error from camera to BEV Polar space due to vehicle pitch variations, imprecise intrinsic or extrinsic camera calibration, labelling method employed and poor synchronization between the camera and radar. Nevertheless, our predictions followed the ground-truth as expected.

Our method could be deployable within autonomous driving systems. However, improper understanding or usage may lead to performance degradation, thereby increasing security risks.

\section{CONCLUSION AND FUTURE WORK} \label{conclusion}
In this work, upon proposing a fusion strategy in BEV space, we analysed how the performance affects the computational metrics in various aspects. Our approach demonstrates proficient performance while upholding a comparatively lower level of computational complexity which align with our research motivation and results shown on RADIal dataset. Nevertheless, obtaining high-quality time synchronized multi-modal data with precise annotations require considerable effort. Hence a further potential direction is to build a diversely recorded large-scale high-quality dataset to accelerate further research.

\bibliographystyle{ieeetr}
\bibliography{references_z.bib}

\end{document}